\documentclass[letterpaper]{article} 
\usepackage{aaai2026}  
\usepackage{times}  
\usepackage{helvet}  
\usepackage{courier}  
\usepackage[hyphens]{url}  
\usepackage{graphicx} 
\usepackage{natbib}  
\usepackage{caption} 
\usepackage{booktabs}
\usepackage{amsmath,amsfonts}
\usepackage{algorithm}
\usepackage{algorithmic}
\usepackage{multirow}

\usepackage{newfloat}
\usepackage{listings}
\DeclareCaptionStyle{ruled}{labelfont=normalfont,labelsep=colon,strut=off} 
\floatstyle{ruled}
\newfloat{listing}{tb}{lst}{}
\floatname{listing}{Listing}
\title{Towards Unified  Vision-Language Models with  Incomplete Multi-Modal Inputs}

\author{
    Xiang Fang\textsuperscript{\rm 1}, Wanlong Fang\textsuperscript{\rm 2}, Changshuo Wang\textsuperscript{\rm 3}\thanks{Corresponding Author. }, Keke Tang\textsuperscript{\rm 4},
    Daizong Liu\textsuperscript{\rm 5},  Siyi Wang\textsuperscript{\rm 2}, Wei Ji\textsuperscript{\rm 6}
}
\affiliations{
    \textsuperscript{\rm 1}School of Software Engineering,
Huazhong University of Science and Technology\\
    \textsuperscript{\rm 2}Nanyang Technological University, Singapore\\ 
    \textsuperscript{\rm 3}University College London \\
    \textsuperscript{\rm 4}Guangzhou University \\
    \textsuperscript{\rm 5}Wuhan University \\
    \textsuperscript{\rm 6}Nanjing University 

     xfang9508@gmail.com,  wanlongfang@gmail.com, wangchangshuo1@gmail.com, tangbohutbh@gmail.com, daizongliu@whu.edu.cn, siyi002@e.ntu.edu.sg, weiji0523@gmail.com
}

\begin{document}
 \maketitle
 \begin{abstract}
Video-Language Models (VLMs) have demonstrated impressive multi-modal reasoning capabilities across diverse computer vision applications. However, these VLMs are task-specific and assume that both video and language inputs are complete. However, real-world VLM applications might face challenges due to deactivated sensors (e.g., cameras are unavailable due to data privacy), yielding modality-incomplete data and leading to inconsistency between training and testing data.  While straightforward incomplete input can boast training generalization-ability and lead to training failure, its potential risks to VLMs regarding safety and trustworthiness have been largely neglected.  To this end, we make the first attempt to propose a unified incomplete video-language model to process the incomplete multi-modal inputs. 
Extensive experimental results show
that our  method can serve as a plug-and-play module for previous works to improve their performance in various multi-modal tasks. 
\end{abstract}    
 \section{Introduction}
\label{sec:intro}
\begin{figure}[t!]
\centering
\includegraphics[width=\columnwidth]{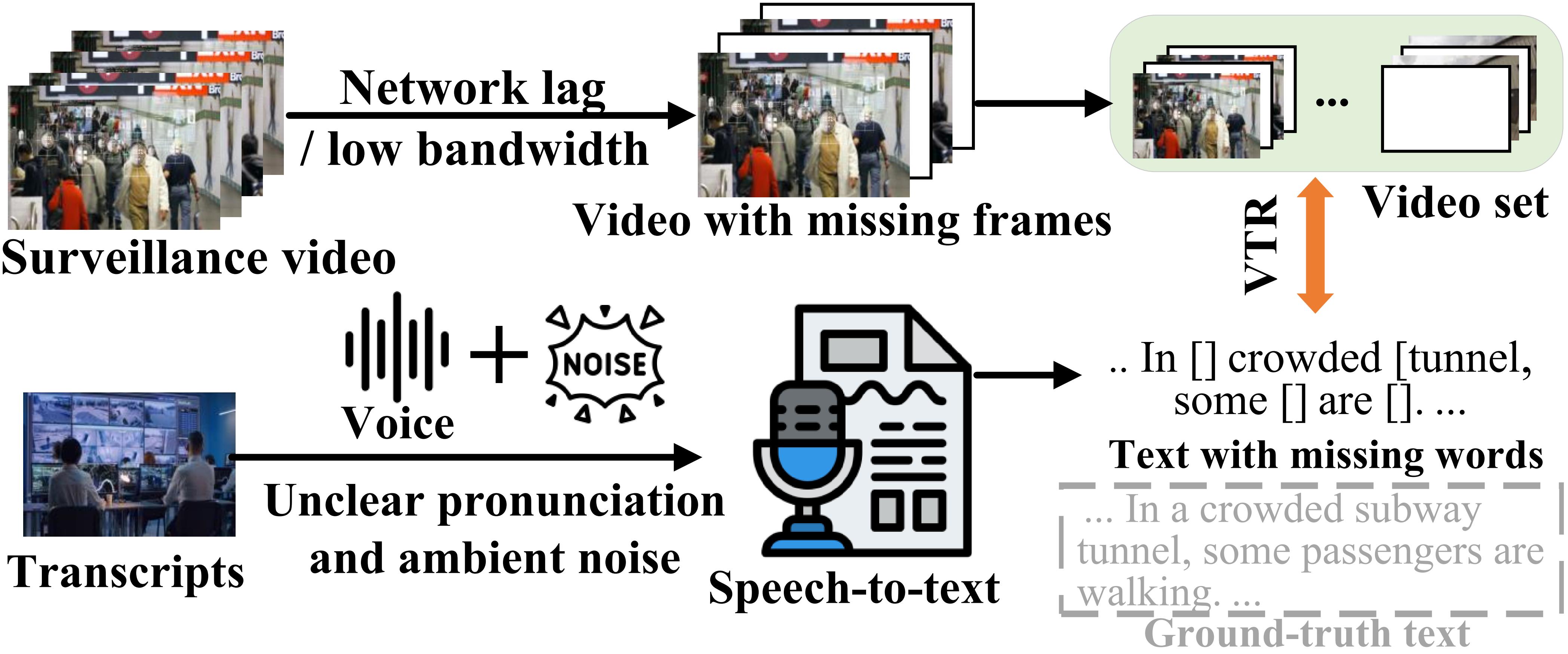}
\caption{Real-world ``surveillance video analysis'' application with missing frames and words for video-text retrieval. }
\label{fig:intro_miss}
\end{figure}

Video-Language Models (VLMs) \cite{momeni2023verbs,fang2025your,fang2023hierarchical,fang2022multi} have achieved significant success and demonstrated promising capabilities in various multi-modal downstream applications, such as text-to-video retrieval \cite{ventura2024learning,fang2023you,fang2025hierarchical,fang2025adaptive} and video question-answering \cite{xiao2024can,fang2020double,liu2023exploring,wang2025taylor,fang2026towardsicml,kuai2026dynamic,wang2025point,fang2025your,zhang2025monoattack,fang2023hierarchical,liu2024towards,yang2025eood,fang2022multi,fang2026cogniVerse,lei2025exploring,fang2023you,wang2025dypolyseg,fang2025hierarchical,yan2026fit,fang2025adaptive,wang2026topadapter,cai2025imperceptible,fang2026slap,wang2026reasoning,fang2026immuno,wang2026biologically,fang2026disentangling,wang2025reducing,fang2026advancing,fang2026unveiling,wang2026from,liu2023conditional,liu2026attacking,fang2026rethinking,wang2025seeing,fang2025multi,fang2024fewer,liu2024pandora,fang2024multi,fang2025turing,fang2024not,liu2023hypotheses,fang2024rethinking,liu2024unsupervised,fang2023annotations,xiong2024rethinking,fang2021unbalanced,wang2025prototype,zhang2025manipulating,fang2026align,tang2024reparameterization,fang2025adaptivetai,tang2025simplification,fang2021animc,cai2026towards,fang2020v}. 
However, with the exponential expansion of downstream applications in the real world, VLMs might face network instability or data loss \cite{bordes2024introduction,wang2025point,wang2025reasoning,wang2025dypolyseg,wang2026biologically,wang2025taylor,li2025meta,ijcai2025p908}, posing incomplete multi-modal inputs \cite{jang2024towards}. In real-world applications, some frames are missing in the videos, while some words are unavailable in the texts/sentences. As shown in Figure \ref{fig:intro_miss}, the ``surveillance video analysis'' application could contain missing frames and words. 1) Missing frames: In security monitoring, cameras may lose frames due to network lag or low bandwidth. 2) Missing words: Transcripts  from operators  may contain missing or unclear words due to noise or overlapping speech. Missing frames may cause crucial actions (\textit{e.g.}, theft) to be lost, making retrieval harder.
Incomplete  texts make it challenging to match key moments accurately based on limited words, leading to a failed retrieval.
Besides,  different modalities have various incompleteness rates\footnote{Definition for  incompleteness rate: incomplete(video) = \#(missing frames)/\#(total frames) and incomplete(text) = \#(missing words)/\#(total words). Balanced incompleteness: incomplete(video) = incomplete(text); unbalanced incompleteness: incomplete(video) $\neq$ incomplete(text).}, which leads to unbalanced incompleteness.

\begin{figure*}[t!]
\centering
\includegraphics[width=\textwidth]{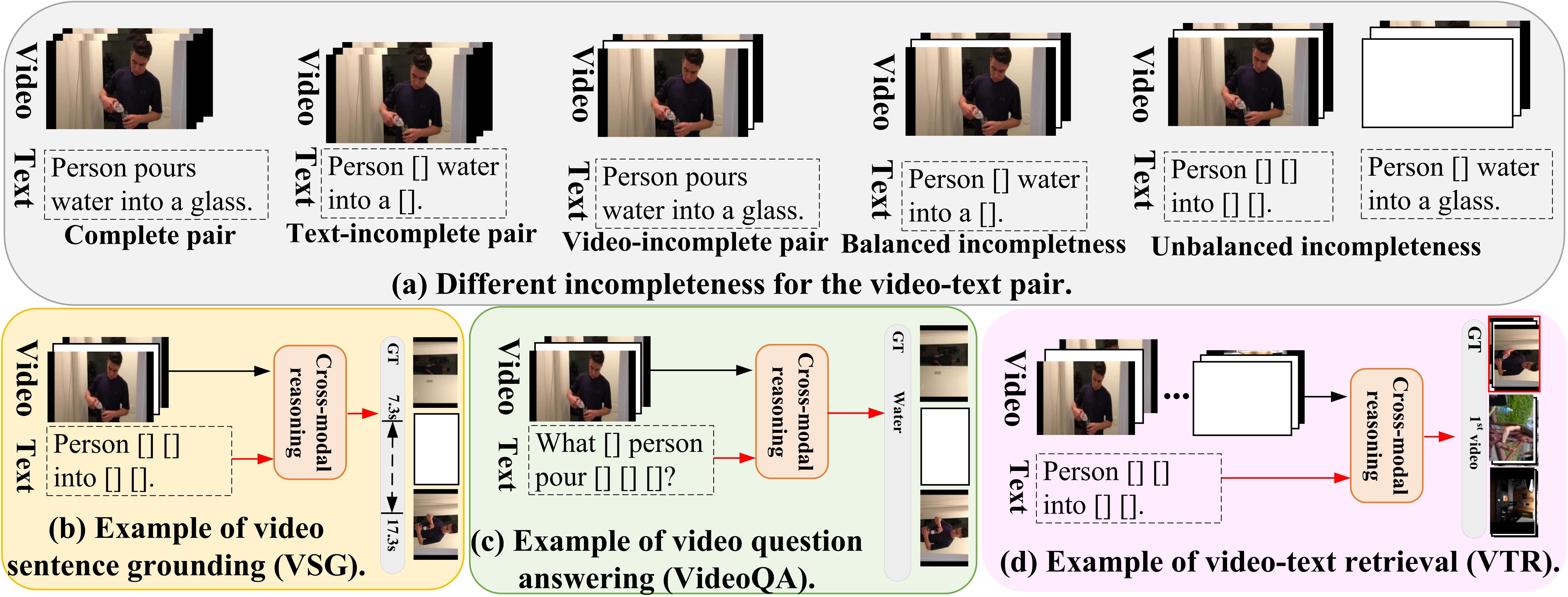}
\caption{Incomplete multi-modal inputs for different multi-modal tasks. (a) Different incompleteness for the video-text pair. Incomplete pair for different tasks: VSG (b), VideoQA (c), VTR (d). ``[]'' means that the corresponding word is missing.}
\label{fig:intro}
\end{figure*}

The downstream VLM-based methods \cite{fang2026align,zhang2025can} focus on the multi-grained information alignment between video and text. Recently, these VLM-based methods have achieved  significant success by
first projecting the video and text features into a common feature space and then introducing a  loss for cross-modal alignment. 
Unfortunately, these VLM-based methods  rely heavily on the complete video-text pairs during training and inference. In fact, during multi-modal data acquisition and processing, data missing and corruption will inevitably occur.
Besides, most methods use uniform or fixed-rate frame sampling to understand motion information in videos. For incomplete video, they cannot reconstruct the correct motion due to missing frames, leading to performance degradation or even model failure.
Consequently, previous methods cannot effectively understand  incomplete videos/texts for cross-modal alignment.


In this paper, we pose a more practical setting called an incomplete video-language alignment, where only incomplete video-text pairs are available during  training and inference.
However, the realistic task faces  the following essential challenges of incomplete multi-modal inputs among various  downstream multi-modal tasks: 1) Existing VLM-based methods severely rely on the complete multi-modal inputs, which limits their applications with incomplete multi-modal inputs since most users do not upload all the information to the target multi-modal applications. In this case, these state-of-the-art methods will suffer severe performance degradation. 2) VLMs have great capabilities of handling multiple vision-language tasks with different prompts. However, existing VLM methods only deceive a specific task. When compromising different downstream tasks, we have to design a distinct multi-modal fusion method, which incurs significant time and resource expenditure. 
To make the attack more robust with high generalization-ability, 
we target to design a unified  completion strategy for various incomplete multi-modal inputs across different  downstream tasks.

To tackle the above issues and increase the robustness of incomplete video-language models for real-world applications. 
To this end, we pose a brand-new setting for VLMs, unbalanced incomplete VLM, where videos and texts are incomplete and have different incompleteness rates. In this work, we define a novel task termed unbalanced incomplete video-language model  and construct many datasets to benchmark the challenging settings in various downstream multi-modal tasks (video-text retrieval, video question answering and video sentence grounding).
To handle the challenging and realistic setting, 
we make the first attempt to explore a task-agnostic modality completion method for different video-language models.
Especially, our proposed framework consists of three modules: multi-modal feature approximation, multi-modal knowledge distillation and multi-granularity  multi-modal integration. Our main contributions are summarized as follows:
1) As far as we know, we make the first attempt to pose a brand-new and realistic setting, incomplete video-text alignment for unbalanced incomplete multi-modal inputs. 
We propose a unified completeness network to address the modality-incomplete challenges in various downstream multi-modal tasks. 2)     We  design a multi-modal feature approximation module, which can approximate more reliable completion features for the incomplete modalities. 
Also, we propose a multi-modal knowledge distillation module to reduce over-reliance on the complete modality.
In the multi-granularity  multi-modal integration module, we integrate semantics-similar video-text pairs by mapping them more compactly in the common feature space.
3)  Extensive experimental results on several benchmarks with different incompleteness rates amply demonstrate that  our proposed method can serve as a plug-and-play module for various state-of-the-art task-specific to improve their performance in various multi-modal tasks.
 \section{Related Works}
\label{sec:related}


\noindent \textbf{Incomplete multi-modal inputs.}
Real-world multi-modal applications always suffer modality incompleteness since the sample collection in some modalities is very labor-intensive and time-consuming \cite{hu2024deep}. Recently, some works focus on improving the model robustness on modality-incomplete data across various multi-modal tasks~\cite{zhao2016incomplete}. Some methods aim to optimize the multi-modal fusion strategy \cite{ma2022multimodal}, while other methods try to conduct data augmentation \cite{mckinzie2023robustness} or regularize objectives \cite{mckinzie2023robustness} to complete the missing samples.
Unfortunately, these methods only perform well on simple classification tasks. When facing some  complex multi-modal tasks (\textit{e.g.}, VSG), they often achieve unsatisfactory performance.


\noindent \textbf{Multi-modal  learning.}
Multi-modal learning leverages multiple types of data (\textit{e.g.}, text,  visual) to create models that better understand complex, multi-faceted information \cite{huang2021makes}. Multi-modal methods \cite{zhu2024unimod1k} have gained traction as they are more aligned with real-world data, which is rarely unimodal.
Due to remarkable success, multi-modal learning has achieved attracted more and more attention \cite{tian2024argue}. State-of-the-art multi-modal methods \cite{sun2024multi} achieve  performance improvement on some tasks under the strict assumption of complete modalities. In many real-world applications, only a subset of modalities are available during training and inference, limiting the performance of these methods.

 \section{Our Proposed Method}
\label{method}

\begin{figure*}[t!]
\centering
\includegraphics[width=\textwidth]{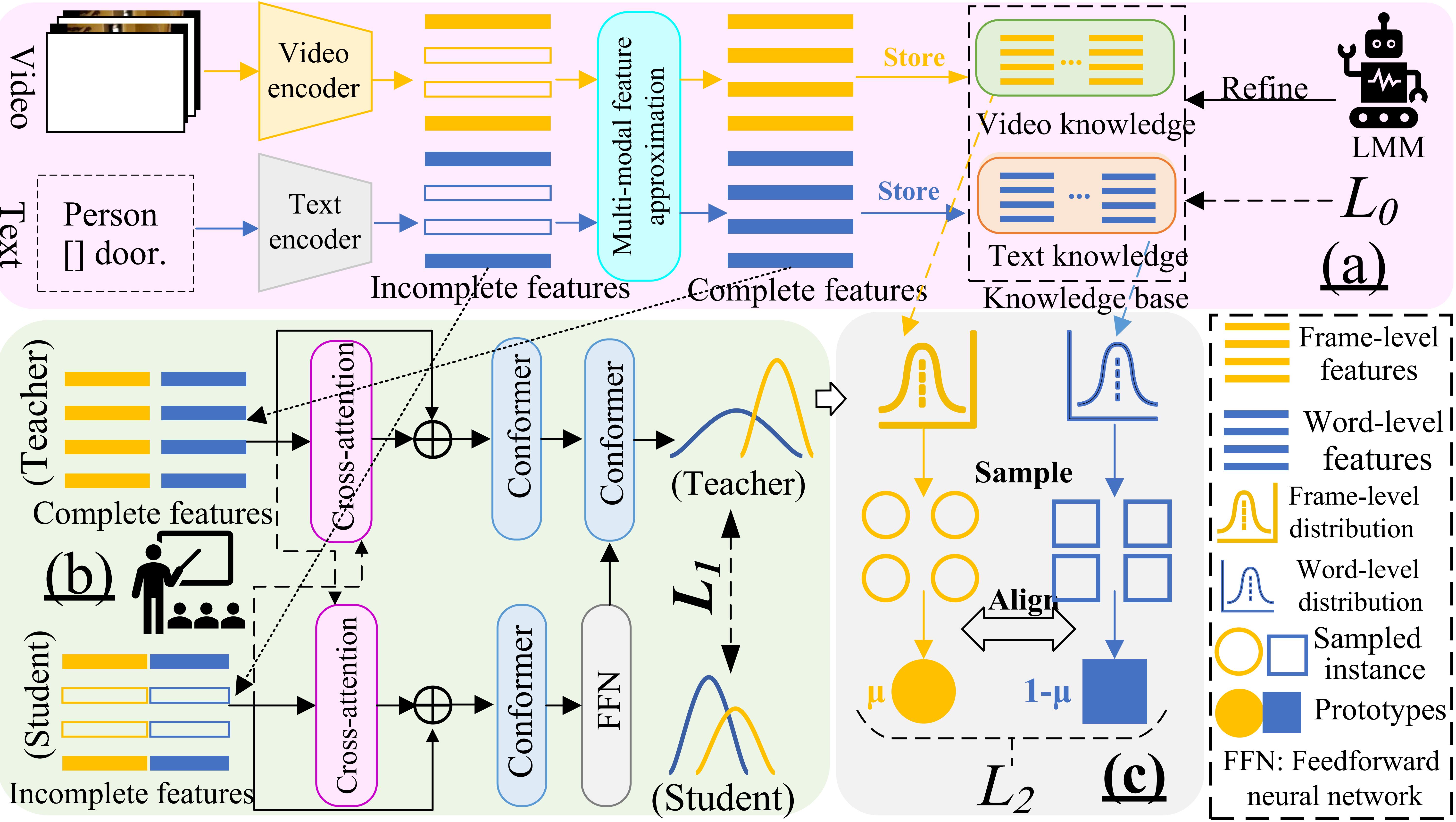}
\caption{Overview of the proposed architecture for the incomplete video-text pair, where (a) is the ``Multi-modal Feature Approximation'' module, (b) is the ``Multi-modal Knowledge Distillation'' module, (c) is the ``Multi-granularity  Multi-modal Integration'' module. 
}
\label{fig:pipeline}
\end{figure*}



\noindent \textbf{Problem statement.}
Given an incomplete multi-modal set $\{\mathbb{V}_n, \mathbb{Q}_n, \mathbb{Y}_n\}_{n=1}^N$, each untrimmed video $\mathbb{V}_n$ is represented as $\mathbb{V}_n=\{v_{n,t}\}^{T}_{t=1}$ frame-by-frame, where $v_{n,t}$ is the $t$-th frame of $n$-th video and $T$ is the number of total frames. Similarly, the sentence text $\mathbb{Q}_n$ with $M$ words is denoted as $\mathbb{Q}_n=\{q_{n,m}\}^{M}_{m=1}$ word-by-word. For missing frame or word, we treat it as null and denote as ``$[]$''. 
Previous VLM-based methods \cite{weng2025longvlm} fail to address the incompleteness challenge since they cannot deal with these incomplete inputs. When directly using these incomplete inputs for the downstream multi-modal tasks, their performance will drop significantly.
To address the above challenges about incomplete multi-modal inputs, we propose a novel plug-and-play multi-modal feature approximation to approximate the missing frame/word features. These approximated features will provide significant semantics for  multi-modal alignment. The proposed approach is application-agnostic and can be adopted successfully in the multi-modal task.

\noindent \textbf{Pipeline.} 
Our pipeline is summarized in Figure~\ref{fig:pipeline}. 
Given an incomplete video-text pair, we first design a multi-modal feature approximation module to construct relational multi-modal graphs based on available cross-modal high semantic similarity features, which can approximate more reliable completion features for the missing modalities.
Then, we propose a multi-modal knowledge distillation module to reduce over-reliance on the complete modality and to balance performance and robustness.
Finally, we propose a prototype-based weighted multi-modal integration module to map semantically similar video-text pairs more compactly in the common embedding space.


\subsection{Multi-modal Feature Approximation}
Given incomplete video-text inputs, we first utilize the feature encoder networks to extract the the visual and textual features, where the video is encoded  frame-by-frame and the text is encoded  word-by-word. For the given incomplete video-text pair $\{V_n,T_n\}_{n=1}^N$, we utilize the feature encoder to obtain the initial features ($v'_i,t'_i$), where $v'_i$ and $t'_i$ denote  video  and  text features, respectively. To conduct the cross-modal integration and semantic alignment, we need to project multi-modal features into a joint feature space. Thus, we introduce the prototype learning strategy to conduct   fine-grained  multi-modal alignment by constructing the shared  prototypes. For convenience, we denote the shared  prototypes across videos and texts as $P \in \mathbb{R}^{N_p\times d}$, where $N$ denotes the total number of video-text pairs, and $N_p$ and $d$ denote the prototype number and the feature dimension, respectively. Firstly, we randomly initialize these prototypes, and then update the prototypes  during training. To align videos and texts for cross-modal fusion, we treat the shared prototype $P$ as the query in the transformer’s cross-attention operation, while the original video features $v_i$ as the key $W_K$ and value $W_V$.
We utilize the video feature as an example, and the same is true for text features. Therefore, we can obtain the reconstructed  features: $v_i=v'_i+FFN(v'_i+MCA(P,v'_i)),t_i=t'_i+FFN(t'_i+MCA(P,t'_i))$, where $v_i$ and $t_i$ denote the corresponding reconstructed features, $MCA(\cdot)$ denotes multi-head cross-attention, and $FFN(\cdot)$ denotes the feed-forward network.

Due to the continuity of the video frames, we can approximate missing frames using features of neighboring frames. Thus, we target to complete the fine-grained features (word  and frame features).
Especially, we first introduce a Jaccard distance function to compute the distance between two nearest neighbor samples. Then, we choose the most reliably $K$-reciprocal nearest neighbors from cross-modality and self-modality. For the missing frame feature $v_a$, we can obtain the semantics-relevant text feature $t_a$. Then, we compute the cosine similarity between $t_a$ and all the frame features $\{v_i\}_{i=1}^T$, where $T$ is the total frame number in the given video. After that, we rank and identify the $K$ most similar frame features to the word feature $t_a$. We denote the $K$ most similar frame features as $N_K(t_a)=\{v_1,...,v_K\}$. 
Similarly, for any $v_i \in N_K(t_a)$, we compute the cosine similarity between $v_i$ and all the word features. Thus, we can have the $K$ most similar word features are denoted as  $N_K(v_i)=\{t_1,...,t_K\}$.
For the word feature $t_a$, we can obtain the  cross-modality $K$-reciprocal nearest neighbors $\mathcal{O}_K(t_a)$ as follows: $\mathcal{O}_K(t_a)=\{v_i|(t_a\in N_K(v_i))\cap (v_i\in N_K(t_a))\}$.
Besides the cross-modal semantics between videos and texts, we can explore the semantic relationship within videos or texts. Therefore, for any two frames $v_i, v_j \in N_K(t_a)$, we  compute intra-modal cosine similarity between $v_i$, $v_j$ and all existing frame features to get the $K$-nearest neighbor sets $N_K(v_i)$ and $N_K(v_j)$ for $v_i$ and $v_j$. In the single-modal setting, the $K$-reciprocal nearest neighbor is computed by: $\mathcal{O}_K(v_i)=\{v_j|(v_j\in N_K(v_i))\cap(v_i\in N_K(v_j))\}$.
Combining multi- and single-modal $K$-reciprocal nearest neighbors, we introduce the following  Jaccard distance $J(t_a,v_i)$ to compute the distance between $t_a$ and $v_i$: $J(v_i,t_a)=\frac{|\mathcal{O}_K(v_i)\cup\mathcal{O}_K(t_a)|-|\mathcal{O}_K(v_i)\cap\mathcal{O}_K(t_a)|}{|\mathcal{O}_K(v_i)\cup\mathcal{O}_K(t_a)|}.$
Thus, to generate more accurate  nearest neighbors, we try to search  ${K_0}$-reciprocal nearest neighbors. Thus, the high semantic similarity neighbor generation set is $N_{{K_0}}(t_a)=\{v_1,v_2,...,v_{{K_0}}\}$.
The same applies to the missing text features as well.

Since the calculation of multi-modal fusion is  feature-based, we complete the incomplete modality from the feature level.
For missing features (frame feature $v_a$ and word feature $t_b$), we can obtain $K$ most relevant nearest neighbor sets for each modality: $N_K(t_a) = \{v_1, v_2, ..., v_{K_0}\}$ and $N_K(v_b) = \{t_1, t_2, ..., t_{K_0}\}$. Thus, we can obtain the approximated features ($v_a$ of $V_a$ and $t_b$ of $T_b$): $\hat{v}_a=M_v\cdot[t_a,N_{K_0}(t_a)],\quad\hat{t}_b=M_t\cdot[v_b,N_{{K_0}}(v_b)],$
where $M_v$ and $M_t$ denote the affinity matrices of $[t_a,N_{K_0}(t_a)] = [t_a, v_1, v_2, ..., v_{K_0}] = [h_1, h_2, ..., h_{K_0+1}]$ and $[v_b,N_{{K_0}}(v_b)] = [v_b, t_1, t_2, ..., t_{K_0}]$. 
Each value denotes a semantic similarity score between two instances (frame or word). Then, we can construct the video memory $\hat{V}_a=\{\hat{v}^1_a,\hat{v}^2_a,...,\hat{v}^{K_0}_a\}$ and the text memory $\hat{T}_b=\{\hat{t}^1_b,\hat{t}^2_b,...,\hat{t}^{K_0}_b\}$. To ensure that our feature approximation strategy is sensible in the real world, we introduce a pre-trained large multi-modal model (LMM) to refine the approximated features: $[{\hat{V}_a,\hat{T}_b}]=LMM([{\hat{V}_a,\hat{T}_b}])$. 
Also, we have $M_v=M_L^{-1} \cdot M_1$,
where $M_L^{-1}$ denotes the normalized Laplacian matrix of $M_1$, and each element $M_{1_{ij}} \in M_1$ is obtained by $M_{1_{ij}}=\exp(\cos(h_i,h_j))$,
where $\cos(\cdot,\cdot)$ denotes the cosine similarity function. Similarly, we can conduct a similar process to $M_t$. 

Please note that the above method is essentially equivalent to constructing graph relationships, where we can transmit information across different samples within the graph and enhance the feature completion. In the graph, the affinity matrix $M_v$ can be treated as the edges and the feature $[t_a,N_K(t_a)]$ serves as the nodes.
Thus, we can formulate our multi-modal feature approximation module as follows:
\small
 \begin{equation} \label{fmc}
\mathcal{L}_0=\frac{1}{N_m^v}\sum\nolimits_{a=1}^{N_m^v}||t_{a}-\hat{v}_{a}||_{2}^{2}+\frac{1}{N_m^t}\sum\nolimits_{a=1}^{N_m^t}||v_{a}-\hat{t}_{a}||_{2}^{2},
\end{equation}\normalsize
where $N_m^v$ and $N_m^t$ denote the numbers of missing frames and missing words, respectively.
By Eq. \eqref{fmc}, we can mitigate the modal discrepancy between the approximated features and the original features.

\subsection{Multi-modal Knowledge Distillation}
Completing all the missing frames and  words by the multi-modal feature approximation module is time-consuming, especially for long videos.
To  reduce over-reliance on the complete modality and to balance performance and robustness,
we design a novel multi-modal knowledge distillation module to train an efficient student model that 
 does not require the expensive feature approximation.


Different from previous knowledge distillation methods, the difference between teacher and student models in our method is the modal gap, not data size. Especially, we  train the teacher model on the complete video-text pairs and the student model on incomplete pairs. Compared with the  student model, the teacher model is relatively unbiased with a higher rate of modality-general decisive features $f^c$ in the shared space. When we train the student model, we treat the teacher model as an anchor point, which can prevent the student model from shifting towards an unimodal distribution in the text modality. In our module, we conduct the knowledge at the hidden layer, not the logistic outputs, which can minimize the distances between the decision distribution samples of the teacher and student models. Besides, our knowledge distillation module constrains the intermediate representation subspace distribution of the student model. Therefore, we can take the knowledge from the intermediate representation of the cross-modal encoder layers for the downstream multi-modal task.


In our model, the samples from original feature space $\mathcal{S}^{v} \times \mathcal{S}^{t} \times \mathcal{Y}$ can be denoted as triples ($v,t,y$). 
For the teacher model, we first train the teacher model $\mathcal{T}(\theta)$ on a complete multi-modal data ($v,t,y$) model with parameters $\theta$. Then, we can obtain the model's decisions ($P_1(y|t, v)$ and $P_2(f^c|t, v_{k_i})$) in a Bayesian decision problem. Since our model is expected as a unified network for multiple downstream tasks, we train the teacher model by minimizing the following loss function for multi-task learning:

\small
\begin{equation}
\mathcal{T}(\theta)=\min _\theta \mathcal{L}_{\text{MLT}}(g(v,t ; \theta), y),
\end{equation}\normalsize
\small
\begin{equation}
\begin{aligned}
\label{loss}
\mathcal{L}_{\text{MLT}}(v,t; \theta) \!=\! \mu \log P_{\text{CTC}}(y|t, v)  \!+\! (1\!-\!\mu) \log P_{\text{Att}}(y_i|t, v),
\end{aligned}
\end{equation}\normalsize
where  $\mu \in (0,1)$ is a parameter to balance different losses.
During  training the student model, we leverage the dropout strategy \cite{mckinzie2023robustness} on the video modality $v$, while we freeze the teacher model with complete video-text pairs as  multi-modal inputs. Please note that the student and teacher models share a similar network architecture. Besides, we divide the whole decision process of the multi-modal model into a hidden feature generation step and a decision step for better interpretability. We have $P_2(y|t, v_{k_i})=P_2(y|f^c) P_2(f^c|t, v_{k_i}),P_1(y|t, v)=P_1(y | f^c) P_1(f^c|t, v),$
where $f^c\in \mathbb{R}^{d}$ denotes the combined feature of modality-specific decisive features $f^{t},f^{v} \in \mathbb{R}^{d}$, and modality-general decisive features $f^c \in \mathbb{R}^{d}$. The tuple $(f^{t}, f^{v}, f^c )$ denotes a sample drawn from the shared features space, which denotes $\mathcal{S}^{t} \times \mathcal{S}^{v} \times \mathcal{S}^{c}$.

During training, besides initializing the parameter of the teacher model, we utilize an additional loss for constraining the dynamic process of the student model's feature distribution. To conduct the frame-level knowledge distillation, we approximate the difference of distribution by the distance between batch samples from the student and  teacher models. The loss is as defined as $\mathcal{L}_{\text{KD}}(v,t, v_{k}) = \text{KL}(
    S_{t}, S_{s})$, where $S_{t} = \delta_{\sigma}(\mathcal{F}_s(P_1(f^c|t, v)))$ and $S_{s}= \delta_{\sigma}(\mathcal{F}_s(P_2(f^c|t, v_{k_i})))$, 
where $\mathcal{F}_s(\cdot)$ and $\delta_{\sigma}(\cdot)$ denote the sample function and the SoftMax function with temperature $\sigma$, respectively. 

Three main purposes are considered in the distribution approximation: 1) by the dual cross-attention design, the process complements the information extracted from $x^a$, which can  effectively address the condition of missing frames and promote out-of-distribution generality. 2)  when the student network encounters a missing modality feature $v_{k_i}$ during training, the convergence of the student's decisive feature $z^u=g(t,v_{k_i};\theta_{s})$ towards the teacher's decisive feature $z^u=g(t,v;\theta_{t})$ encourages the utilization of contextual information from $v_{k_i}$. 3)  we can utilize the KD loss to maximize the similarity between the distributions of the teacher and student models, which can prevent the student model from converging to trivial solutions. Finally, we train the student model jointly with a weighted sum of the standard training loss and distillation loss:
\small
\begin{multline}
\mathcal{L}_1(v,t, x^v_{k}) = \beta \mathcal{L}_{\text{KD}}(v,t, x^v_{k}) 
+ (1-\beta) \mathcal{L}_{\text{MLT}}(v,t_k).
\end{multline}\normalsize


\subsection{Multi-granularity  Multi-modal Integration}
In real-world multi-modal applications, there are various data granularities in different tasks. For example, we need fine-grained multi-modal understanding for the video sentence grounding task since we need to localize the target segment based on the language sentence. Unlikely, we only need the coarse-grained multi-modal understanding for the video text retrieval since we can directly finish the retrieval based on the global video and text features. To handle the multi-granularity multi-modal inputs, we design a Multi-granularity  Multi-modal Integration module. Especially, we  introduce different weights based on the matching probability between different instances, and then adjust the frame-word alignment  in the shared space. Also, an empirical observation is that noun phrases consistently share either the same or synonymous attributes within two textual descriptions from the same video. Given the sentence $T_i$, we  extract relevant noun phrases as $P(T_i)=Z_i=\{z_1,z_2,...,z_{N_p}\}$,
where $P$ and $N_p$ denote the noun phrase extractor and the number of noun phrases, respectively.
Besides, we can calculate matching probability weights between instance $i$ and instance $j$ as follows: $W_{i,j}=\frac{{|Z_i\cap Z_j|}/{|Z_i\cup Z_j|}}{\sum_{k=1}^N{|Z_i\cap Z_k|}/{|Z_i\cup Z_k|}},$
where $|Z_i \cap Z_j |$ denotes the count of synonymous noun phrases shared between $Z_i$ and $Z_j$. Similarly,  $|Z_i \cup Z_j |$ denotes the number of noun phrases in the union between $Z_i$ and $Z_j$.
By introducing the weight $\alpha$ to balance the cross-modal alignment of different samples, we can obtain:
\small
\begin{align}
\mathcal{L}_2^{v2t}&\!=\frac{1}{N}\sum\nolimits_{i=1}^{N}\sum\nolimits_{j=1}^{N}\! L(v_{i},t_{j})\cdot (\alpha W_{i,j} \!+\!(1\!-\!\alpha)I_{i,j}),\\
\mathcal{L}_2^{t2v}&\!=\frac{1}{N}\sum\nolimits_{i=1}^{N}\sum\nolimits_{j=1}^{N}\! L(t_{i},v_{j}) \cdot (\alpha W_{i,j}\!+\!(1\!-\!\alpha) I_{i,j}),
\end{align}\normalsize
where  $L(v_{i},t_{j})=-log\frac{exp( \cos(v_{i},t_{j})/\sigma)}{\sum_{k=1}^{N}exp( \cos(v_{i},t_{k})/\sigma)}$ and $L(v_{i},t_{j})=-log\frac{exp( \cos(v_{i},t_{j})/\sigma)}{\sum_{k=1}^{N}exp( \cos(v_{i},t_{k})/\sigma)}$, and
$\alpha \in [0, 1]$ denotes the prior probability that frame $v_i$ is matched with its paired text $t_j$. When $\alpha = 1$, we need to utilize the one-hot labels $I_{ij}$ for cross-modal contrastive learning. However, to better align unpaired text feature $t_j$ with frame feature $v_i$, $\alpha W_{i,j}$ supervises the unpaired samples, while 
$(1-\alpha)I_{i,j}$ provides supervision for paired frame-text samples.
Finally, we can obtain the final loss:
\small
\begin{equation}
\mathcal{L}_2=\mu\mathcal{L}_2^{v2t}+(1-\mu)\mathcal{L}_2^{t2v},
\end{equation}\normalsize
where $\mu$ is a parameter to balance the importance between different modalities.

Thus, our model is trained by the following loss:
\small
\begin{equation}
\mathcal{L}=\mathcal{L}_0+\alpha_1 \mathcal{L}_1+\alpha_2 \mathcal{L}_2,
\end{equation}\normalsize
where $\alpha_1$ and $\alpha_2$ are parameters to balance the significance between different losses.

 \section{Experiment}
\label{exp}

\begin{table}[t!]
\scriptsize
\begin{center}
\setlength{\tabcolsep}{0.3mm}{
\begin{tabular}{c|ccc|ccccccc} 
    \hline 
     \multirow{2}{*}{Method} & \multicolumn{3}{c|}{Complete video-text pair } & \multicolumn{3}{c}{Incomplete video-text pair} \\
     \cline{2-7}
     & R@1\scriptsize{$\uparrow$} & R@5\scriptsize{$\uparrow$} & R@10\scriptsize{$\uparrow$} & R@1\scriptsize{$\uparrow$} & R@5\scriptsize{$\uparrow$} & R@10\scriptsize{$\uparrow$}  \\
    \hline
    \multicolumn{11}{c}{Text-to-video retrieval}\\
    \hline
    {{\textit{\scriptsize{CLIP-ViT-B/32}}}} & & & & & & & & & & \\ 
    X-Pool~\citep{gorti2022x}  & 46.9 & 72.8 & 82.2 &  27.5& 43.6& 50.8 \\
    \textbf{+Ours } &  \textbf{48.1}& \textbf{73.6}& \textbf{84.0}&  \textbf{36.3}& \textbf{63.7}& \textbf{70.4} \\\hline
    DiffusionRet~\citep{jin2023diffusionret}   & 49.0 & 75.2 & 82.7 &  27.7& 44.0& 51.2 \\
    \textbf{+Ours } & \textbf{50.8}& \textbf{78.2}& \textbf{86.3}& \textbf{36.8}& \textbf{64.2}& \textbf{69.5} \\\hline
    CLIP-ViP~\citep{xue2022clip} & 50.1 & 74.8 & 84.6  & 31.4& 44.7& 52.0 \\
    \textbf{+Ours } & \textbf{52.3}& \textbf{77.1}& \textbf{86.0}&  \textbf{41.2}& \textbf{69.5}& \textbf{73.1} \\\hline
    T-MASS \citep{wang2024text}  &{50.2} &75.3
    &{85.1}  & 30.8& 45.1& 52.4 \\
    \textbf{+Ours } & \textbf{51.9}& \textbf{76.8}& \textbf{86.3}& \textbf{42.6}& \textbf{70.3}& \textbf{72.5}  \\
    \hline \hline 
    {\textit{\scriptsize{CLIP-ViT-B/16}}} & & & & & & & & & & \\
    X-Pool~\citep{gorti2022x} &48.2 &73.7 &82.6  & 28.0& 44.2& 51.3 \\
    \textbf{+Ours } &  \textbf{50.2}& \textbf{75.1}& \textbf{84.9}& \textbf{37.4}& \textbf{65.8}& \textbf{72.7} \\\hline
    CLIP-ViP~\citep{xue2022clip} & {54.2} & {77.2} & 84.8  & 28.2& 44.7& 51.5  \\
    \textbf{+Ours } & \textbf{55.9}& \textbf{79.8}& \textbf{86.3}&  \textbf{38.2}& \textbf{66.3}& \textbf{73.4}\\\hline
    T-MASS \citep{wang2024text}  &52.7 &77.1 &{85.6}  &  28.5& 45.3& 52.4  \\
    \textbf{+Ours } & \textbf{53.8}& \textbf{80.5}& \textbf{86.9}&  \textbf{43.6}& \textbf{71.3}&\textbf{ 74.9}\\\hline
    \hline
        \multicolumn{11}{c}{Video-to-text retrieval}\\
    \hline
     {\scriptsize\textit{CLIP-ViT-B/32}} & & & & & \\
    X-Pool~\citep{gorti2022x} & 44.4 & 73.3 & 84.0  & 23.3& 42.8& 50.1 \\
    \textbf{+Ours } & \textbf{46.3}& \textbf{75.9}& \textbf{86.2}&  \textbf{35.0}& \textbf{62.4}& \textbf{71.2} \\ \hline 
    UATVR~\citep{fang2023uatvr} & 46.9 & 73.8 & 83.8 &  26.4& 43.8&51.7  \\
    \textbf{+Ours } &  \textbf{48.0}& \textbf{77.2}& \textbf{85.9} & \textbf{36.6}& \textbf{63.7}&\textbf{ 72.4} \\ \hline 
    T-MASS \citep{wang2024text}  &{47.7} &{78.0} &{86.3} & 29.7& 45.4& 52.9\\ 
    \textbf{+Ours } & \textbf{ 51.2}& \textbf{80.3}& \textbf{88.2} & \textbf{38.7}& \textbf{65.2}& \textbf{73.6} \\ \hline 
    \hline 
    {\scriptsize\textit{CLIP-ViT-B/16}} & & & & & \\
    X-Pool~\citep{gorti2022x} & 46.4 & 73.9  & 84.1 &  26.3& 43.5& 51.7  \\
    \textbf{+Ours } & \textbf{48.9}& \textbf{76.2}& \textbf{87.5} & \textbf{36.8}& \textbf{64.7}& \textbf{75.2} \\ \hline 
     UATVR~\citep{fang2023uatvr} & 48.1& 76.3&85.4  & 26.2& 44.3& 52.0  \\
     \textbf{+Ours } & \textbf{49.2}& \textbf{78.0}& \textbf{88.9}&  \textbf{37.4}& \textbf{66.9}& \textbf{74.0} \\ \hline
    T-MASS \citep{wang2024text} &{50.9} &{80.2} &{88.0}  & 28.5& 44.7& 52.9 \\ 
     \textbf{+Ours } & \textbf{51.8}& \textbf{83.4}& \textbf{92.3}& \textbf{41.0}& \textbf{69.5}& \textbf{75.8}\\ \hline
\end{tabular}}
\end{center}
\caption{Video text retrieval comparisons on  MSR-VTT.
} 
\label{tab:vtr}
\end{table}

\begin{table}[t!]
\scriptsize
\centering
\setlength{\tabcolsep}{1mm}{
\begin{tabular}{cc|ccc|ccccccccc}
\hline
\multirow{2}*{Method} & \multirow{2}*{\# Frames} &\multicolumn{3}{|c|}{Complete  pair}& \multicolumn{3}{c}{Incomplete  pair} \\\cline{3-8}
~ & ~ & Tem & Cau & Des & Tem & Cau & Des \\
\hline 
All-in-One~\citep{wang2023all}   & 32  & 48.6 & 48.0 & 63.2 & 29.8& 31.3& 41.7   \\
\textbf{+Ours} & \textbf{32}& \textbf{51.2}& \textbf{52.9}& \textbf{65.0} & \textbf{38.9}& \textbf{40.3}& \textbf{52.6}  \\\hline 
MIST~\citep{gao2023mist}    & 32 &  56.6 & 54.6 &  66.9 & 31.4& 32.9& 43.7 \\
\textbf{+Ours} & \textbf{32}& \textbf{58.7}& \textbf{57.2}&\textbf{ 68.9}& \textbf{40.3}& \textbf{43.0}& \textbf{54.7} \\\hline
HiTeA~\citep{ye2022hitea}   & 16 & 58.3  & 62.4 &  75.6 & 32.0& 33.4& 43.1 \\
\textbf{+Ours} &  \textbf{16} & \textbf{61.3}& \textbf{66.0}& \textbf{78.2} & \textbf{41.8}& \textbf{44.5}& \textbf{55.3}\\\hline
InternVideo~\citep{wang2022internvideo}  & 8 &  58.5 & 62.5 &  75.8 & 33.1& 34.2& 45.9 \\
\textbf{+Ours} & \textbf{8} & \textbf{63.0} & \textbf{64.8}& \textbf{78.9} & \textbf{42.5}& \textbf{46.3}& \textbf{57.0} \\\hline
BLIP-2~\citep{li2023blip}  & 4 & 67.2  & 70.3 &  79.8 & 36.8& 39.7& 49.6 \\
\textbf{+Ours} & \textbf{4} & \textbf{71.2}& \textbf{73.5}& \textbf{82.3 } & \textbf{49.0}& \textbf{53.1}& \textbf{62.4} \\\hline
\end{tabular}
 }
 \caption{VideoQA performance comparison on NExT-QA.
}\label{tab:nextqa}
\end{table}

\begin{table}[t!]
\centering
\resizebox{0.48\textwidth}{!}{
\setlength{\tabcolsep}{1mm}{
\begin{tabular}{c|cccc|cccccccc}
\hline
 \multirow{2}*{Method (\# Frames)} &\multicolumn{4}{|c|}{Complete  pair}& \multicolumn{4}{c}{Incomplete  pair}\\\cline{2-9}
~& Int & Seq & Pre & Fea & Int & Seq & Pre & Fea \\
\hline 
All-in-One~\citep{wang2023all} (32)   & 47.5 & 50.8 & 47.7 &  44.0 & 25.3& 30.2& 28.4& 23.7 \\
\textbf{+Ours (32)} & \textbf{49.1}& \textbf{52.3}& \textbf{48.5}& \textbf{47.5} & \textbf{32.7}& \textbf{40.8}& \textbf{37.6}& \textbf{34.9} \\\hline
MIST~\citep{gao2023mist} (32)   &  55.5 &  54.2 & 54.2  & 44.4 & 30.4& 35.7& 31.0& 32.3  \\
\textbf{+Ours (32)} &\textbf{ 57.5}& \textbf{58.2}& \textbf{59.3}& \textbf{48.7} & \textbf{38.6}& \textbf{43.1}& \textbf{39.6}& \textbf{38.4} \\\hline
InternVideo~\citep{wang2022internvideo} (8)   &   62.7 &  65.6 &  54.9  &  51.9 &  35.2& 37.9& 33.0& 35.8 \\
\textbf{+Ours (8)} &\textbf{ 64.2}& \textbf{68.1}& \textbf{57.4}& \textbf{56.3} & \textbf{43.1}& \textbf{49.6}&\textbf{ 43.2}& \textbf{48.1} \\\hline
SeViLA~\citep{yu2023self} (4)   &  63.7 & {70.4} &  {63.1}  &  {62.4} & 35.9& 38.1& 32.4& 34.9 \\
\textbf{+Ours (4)} & \textbf{65.2} & \textbf{73.0}& \textbf{65.8}& \textbf{65.7} & \textbf{44.2}& \textbf{50.9}& \textbf{45.8}& \textbf{50.7} \\\hline
BLIP-2~\citep{li2023blip} (4)   &  {65.4}  &  69.0  &   59.7  &  54.2 & 36.7& 39.2& 41.4& 37.5   \\
\textbf{+Ours (4)} & \textbf{68.1}& \textbf{73.1}& \textbf{60.8}& \textbf{57.4} & \textbf{46.3}& \textbf{52.7}& \textbf{46.9}& \textbf{53.2}\\\hline
\end{tabular}}}
 \caption{{Comparison Results on STAR VideoQA dataset, where ``Int'' is ``Interaction'', ``Seq'' is ``Sequence'', ``Pre'' is ``Prediction'', and  ``Fea'' is ``Feasibility''.} 
}\label{tab:star}
\end{table}
\noindent \textbf{Datasets.} For a fair comparison, we use the following open-source video-language datasets to evaluate the effectiveness of our proposed framework in various tasks.
1) For the VTR task, we adopt two datasets: {MSRVTT}~\cite{xu2016msr} and {LSMDC} \cite{liu2019use}.
2) For the VSG  task, we utilize three datasets: ActivityNet Captions \citep{caba2015activitynet}, and Charades-STA \citep{sigurdsson2016hollywood} and {TACoS} \citep{regneri2013grounding}.  3) For the VideoQA task, we use two datasets: {NExT-QA} \cite{xiao2021next} and {STAR} \cite{wu2021star}. Unless otherwise specified in this paper, we default that incomplete pairs refer to a 30\% missing rate (\textit{i.e.}, incomplete(video) = incomplete(text) = 30\%). 

\noindent \textbf{Evaluation metrics.}
For the VTR task, we utilize Recall at rank $\{1,5,10\}$ (R@1, R@5, and R@10) for evaluating the retrieval performance. 
For the VSG task, we evaluate the grounding performance by ``R@n, IoU=m'', which means the percentage of queries having at least one result whose Intersection over Union (IoU) with ground truth is larger than m. We use $n \in \{1,5\}$ for all datasets, $m \in \{0.5,0.7\}$ for ActivityNet Captions and Charades-STA, $m \in \{0.3,0.5\}$ for TACoS.
As for the VideoQA task, we introduce the following metrics: temporal (Tem), causal (Cau), description (Des), interaction (Int), sequence (Seq), prediction (Pre) and feasibility (Fea). 
Bold value denotes the best performance.


\begin{table}[t!]
\scriptsize
    \centering
\setlength{\tabcolsep}{0.4mm}{
    \begin{tabular}{c|c|cccc|ccccccc}
    \hline
    \multirow{3}*{Method} & \multirow{3}*{Type} & \multicolumn{4}{c|}{Complete video-text pair} & \multicolumn{4}{c}{Incomplete video-text pair}\\\cline{3-10}
    ~&~& R@1, & R@1, & R@5, & R@5, & R@1, & R@1, & R@5, & R@5,  \\ 
    ~ & ~ & IoU=0.3 & IoU=0.5 & IoU=0.3 & IoU=0.5& IoU=0.3 & IoU=0.5 & IoU=0.3 & IoU=0.5\\ \hline 
    \multicolumn{10}{c}{ActivityNet Captions}\\ \hline
MMN &FS&65.05&48.59&87.25&79.50& 37.90& 28.53& 60.34& 47.98  \\
\textbf{+Ours}& \textbf{FS}& \textbf{67.32}& \textbf{50.28}&\textbf{ 90.34}& \textbf{80.75}&
\textbf{48.93}& \textbf{43.62}& \textbf{75.57}& \textbf{68.49}\\\hline
G2L  &FS & -&51.68& -& 81.32&  38.12& 31.60& 61.49& 49.88   \\
\textbf{+Ours} & \textbf{FS} &  \textbf{68.57} & \textbf{53.20} & \textbf{91.24} & \textbf{83.72} & \textbf{49.31}&  \textbf{44.82}& \textbf{77.28}&\textbf{69.92} \\
\hline
VCA & WS & 50.45 & 31.00 & 71.79 & 53.83 & 27.34& 18.33& 46.52& 32.40 \\
\textbf{+Ours} & \textbf{WS} &  \textbf{52.83}& \textbf{34.76}& \textbf{73.94}& \textbf{56.11}&  \textbf{35.82}& \textbf{25.47}& \textbf{54.38}& \textbf{40.96}\\\hline
WSTAN & WS & 52.45 & 30.01 & 79.38 & 63.42& 28.11& 19.04& 46.70& 33.97\\
\textbf{+Ours} & \textbf{WS} & \textbf{53.86}&\textbf{ 32.19}& \textbf{81.72}& \textbf{66.31} & \textbf{36.95}& \textbf{26.70}& \textbf{56.42}& \textbf{43.07}\\\hline
CNM &WS&55.68&33.33&-&-& 28.99& 21.34& 48.72& 34.20\\
\textbf{+Ours} & \textbf{WS} &\textbf{ 57.35}& \textbf{35.04}& \textbf{82.96}& \textbf{68.43}& \textbf{38.52}& \textbf{28.99}& \textbf{58.43}& \textbf{46.00}\\\hline
    \multicolumn{10}{c}{Charades-STA}\\ \hline
MMN &FS& 47.31& 27.28& 83.74& 58.41 & 21.03& 14.20& 55.42& 22.87\\
\textbf{+Ours}& \textbf{FS}& \textbf{48.92}& \textbf{28.93}& \textbf{86.73}& \textbf{59.67} & \textbf{29.34}& \textbf{19.72}& \textbf{68.91}& \textbf{34.82}\\\hline
G2L &FS & 47.91& 28.42& 84.80& 59.33  & 22.43& 13.87& 56.20& 22.34  \\
\textbf{+Ours} & \textbf{FS} & \textbf{50.82}& \textbf{31.27} & \textbf{86.95} & \textbf{61.83}  &  \textbf{30.84}& \textbf{20.39}& \textbf{71.15}& \textbf{36.82} \\
\hline
WSTAN & WS &29.35& 12.28& 76.13& 41.53 & 13.83& 7.90& 42.09& 13.08 \\
\textbf{+Ours} & \textbf{WS} & \textbf{31.06}& \textbf{14.29}& \textbf{78.13}& \textbf{42.80}  & \textbf{20.65}& \textbf{9.34}& \textbf{54.26}& \textbf{25.67}\\\hline
CNM &WS& 35.15& 14.95& -& -&  15.29& 8.14& 44.19& 14.27\\
\textbf{+Ours} & \textbf{WS} & \textbf{36.87}& \textbf{16.95}& \textbf{78.66}& \textbf{40.15} & \textbf{22.96}& \textbf{11.43}& \textbf{54.88}& \textbf{27.39}\\\hline
VCA & WS &38.13& 19.57& 78.75& 37.75&  16.24& 9.25& 48.10& 16.73  \\
\textbf{+Ours} & \textbf{WS} & \textbf{41.16}& \textbf{21.06}& \textbf{79.52}& \textbf{40.82}  & \textbf{ 25.39}& \textbf{14.72}& \textbf{ 56.30}& \textbf{29.37}\\\hline
 \end{tabular}}
       \caption{Performance comparison for VSG.}
    \label{tab:vsg_act_cha}
\end{table}

\begin{figure}[t!]
\centering
\includegraphics[width=\columnwidth]{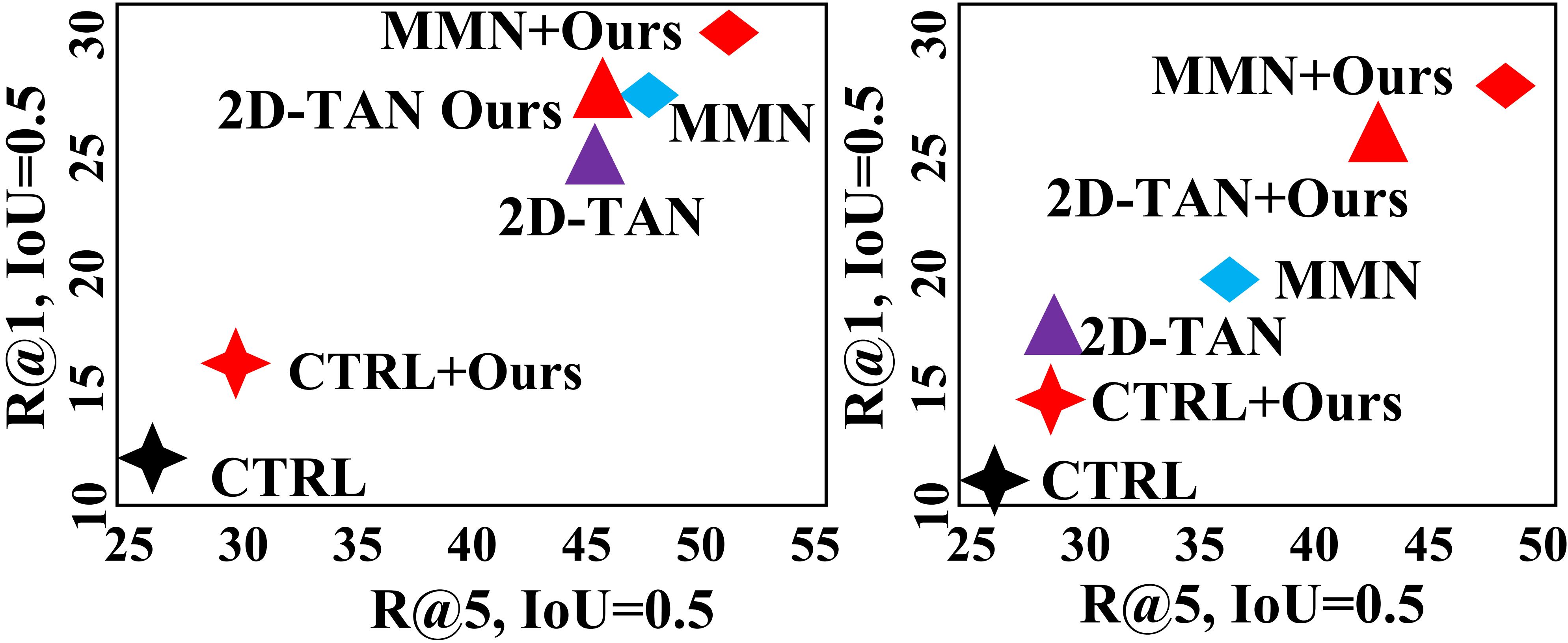}
\caption{VSG performance on TACoS, where the left one is complete pair (incompleteness rate is 0\%) and the right one is balanced incomplete pair (incompleteness rate is 50\%).}
\label{fig:tacos}
\end{figure}

\begin{table}[t!]
\scriptsize
  \setlength{\tabcolsep}{2.5mm}{
    \begin{tabular}{cccc}
    \hline 
    Method & Run-Time & Model Size & R@1, IoU=0.5 \\ \hline
    ACRN \cite{liu2018attentive} & 5.96s & 128M & 13.27 \\
    CTRL \cite{gao2017tall} & 3.58s & \textbf{22M} & 12.13  \\ 
    TGN \cite{chen2018temporally} & 0.89s & 166M & 15.82 \\
    2D-TAN \cite{zhang2019learning} & 0.71s & 232M & 19.96 \\ 
     MomentDiff~\cite{li2023momentdiff} & 1.85s & 248M & 21.40 \\
    \hline
    \textbf{Ours+2D-TAN} & \textbf{0.63s} & 103M & \textbf{26.76} \\ \hline
    \end{tabular}}
        \caption{Efficiency comparison for VSG  on  TACoS.}
	\centering
    \label{tab:efficient}
\end{table}

\begin{table}[t!]
\scriptsize
\setlength{\tabcolsep}{0.48mm}{
\begin{tabular}{c|cccc|cccccccccccccccc}
\hline
\multirow{3}*{Model}&\multicolumn{4}{c|}{ActivityNet Captions} & \multicolumn{4}{c}{Charades-STA} \\\cline{2-9}
& R@1 & R@1 & R@5 & R@5& R@1 & R@1 & R@5 & R@5\\
 & IoU=0.3 & IoU=0.5 & IoU=0.3 & IoU=0.5& IoU=0.5 & IoU=0.7 & IoU=0.5 & IoU=0.7\\\hline
Ours(a)& 40.85& 40.17& 65.93& 60.47& 21.16 & 13.05& 62.40&  28.96 \\
Ours(b) & 44.99& 42.83& 68.19& 65.30& 23.76& 16.37& 67.12& 32.59 \\
Ours(c)& 46.25& 43.72& 70.95& 68.45& 25.73& 18.42& 68.35& 34.80 \\
\hline
\textbf{Ours(full)} &  \textbf{49.31}&  \textbf{44.82}& \textbf{77.28}&\textbf{69.92} & \textbf{30.84}& \textbf{20.39}& \textbf{71.15}& \textbf{36.82} \\ \hline
\end{tabular}}
\caption{Main ablation study for the VSG task with G2L as the base model, where we remove each key individual component to investigate its effectiveness. 
}
\label{tab:main_ablation}
\end{table}

\begin{figure*}[t!]
\centering
\includegraphics[width=\textwidth]{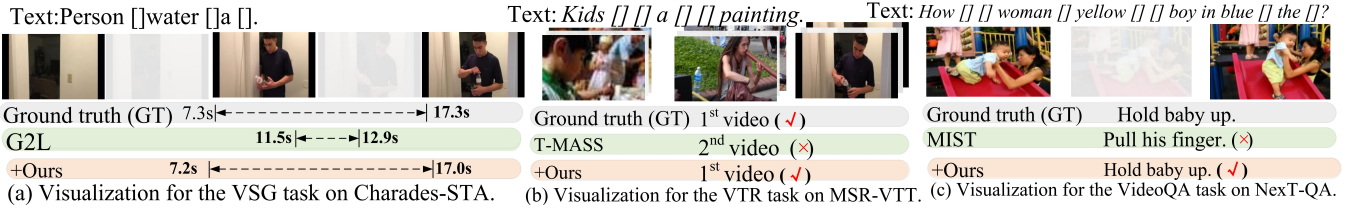}
\caption{Visualization results for different downstream tasks on incomplete multi-modal datasets.}
\label{fig:vis}
\end{figure*}

\begin{figure}[t!]
\centering
\includegraphics[width=0.23\textwidth]{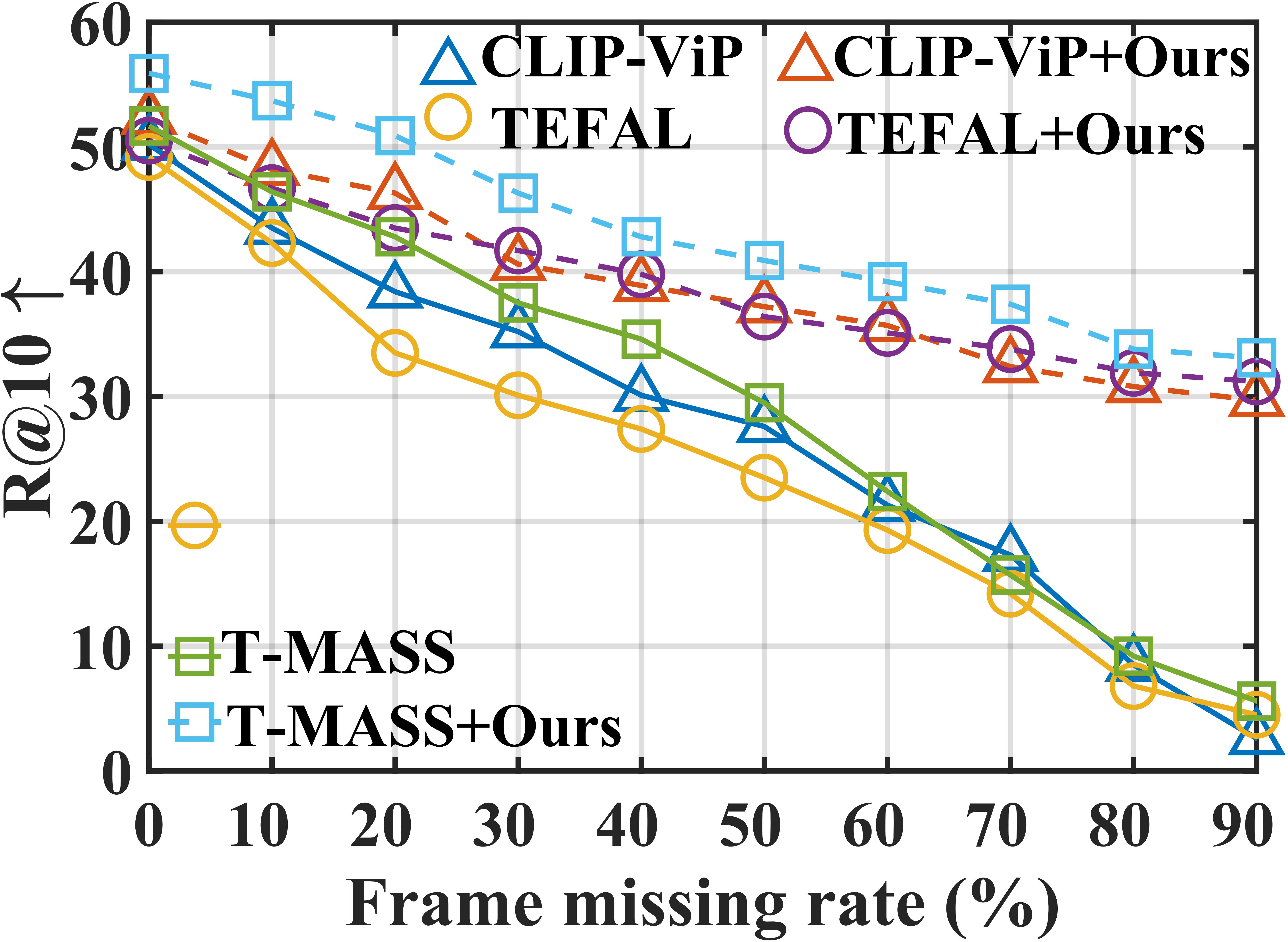} 
\hspace{-0.08in}
\includegraphics[width=0.23\textwidth]{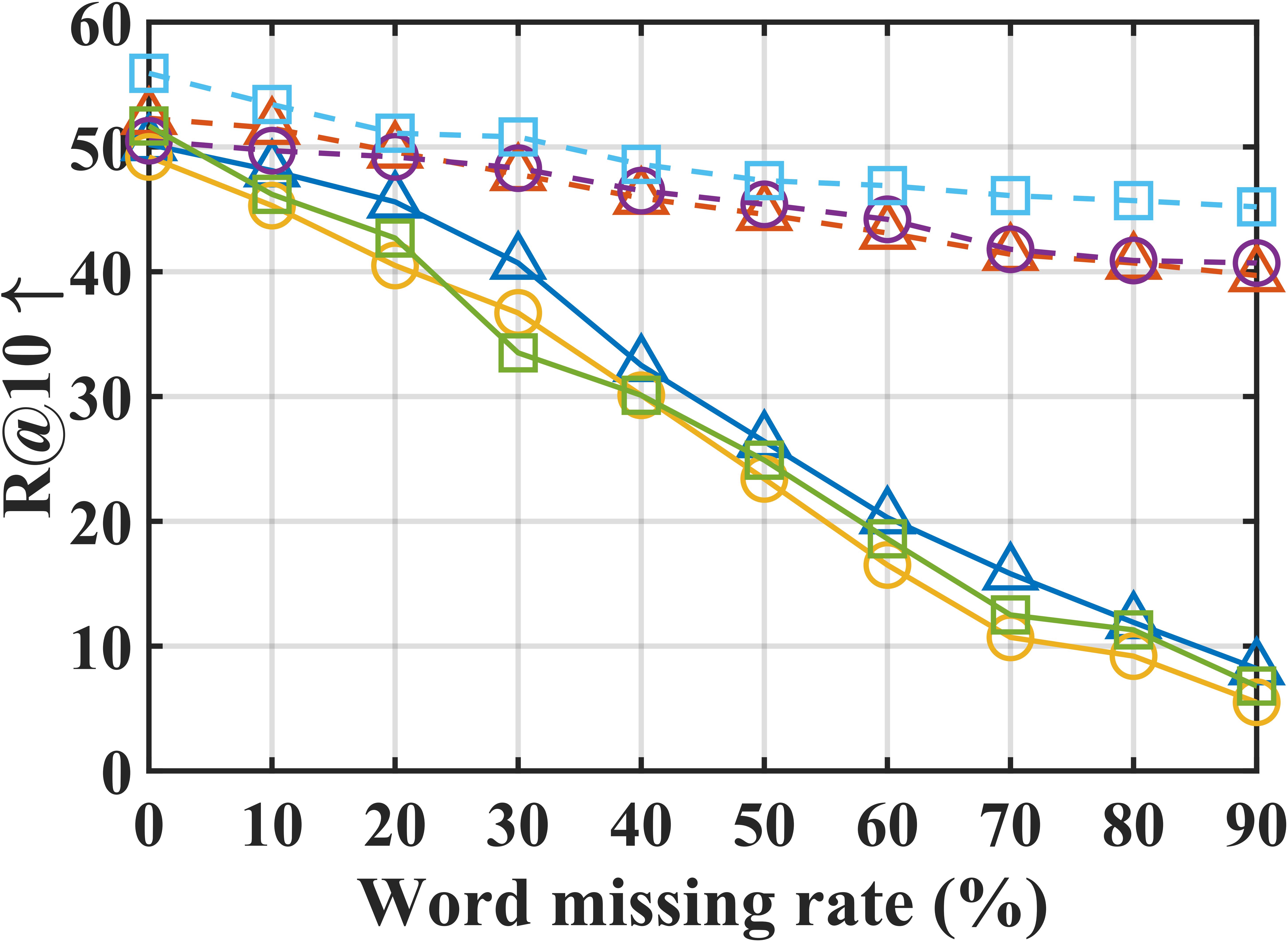} 
\caption{Different incompleteness rates for different modality for the text-to-video retrieval task on the LSMDC dataset (left: incomplete video  and right: incomplete text).}
\label{fig:missing_rate}
\end{figure}

\begin{figure}[t!]
\centering
\includegraphics[width=0.16\textwidth]{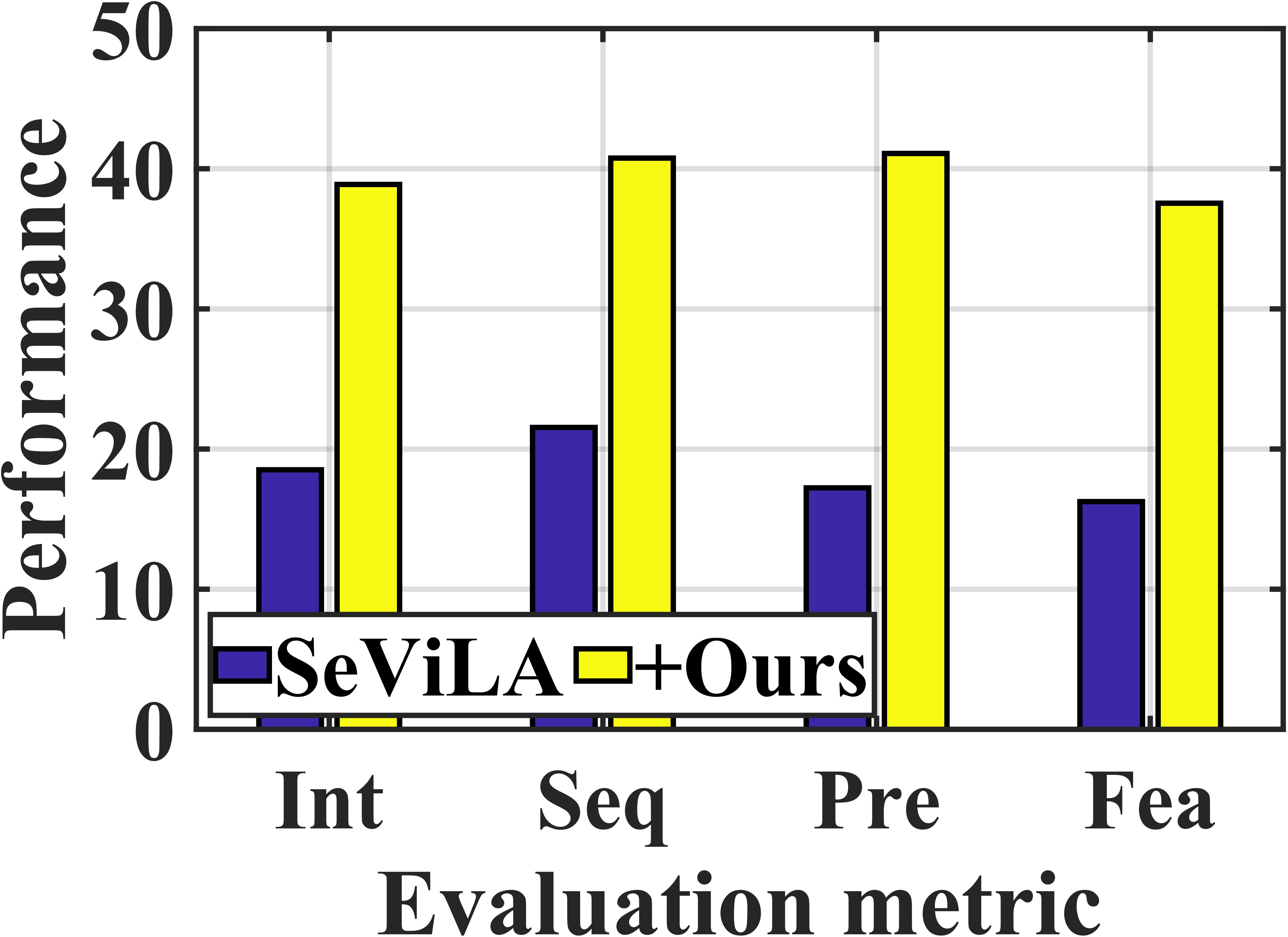} 
\hspace{-0.09in}
\includegraphics[width=0.16\textwidth]{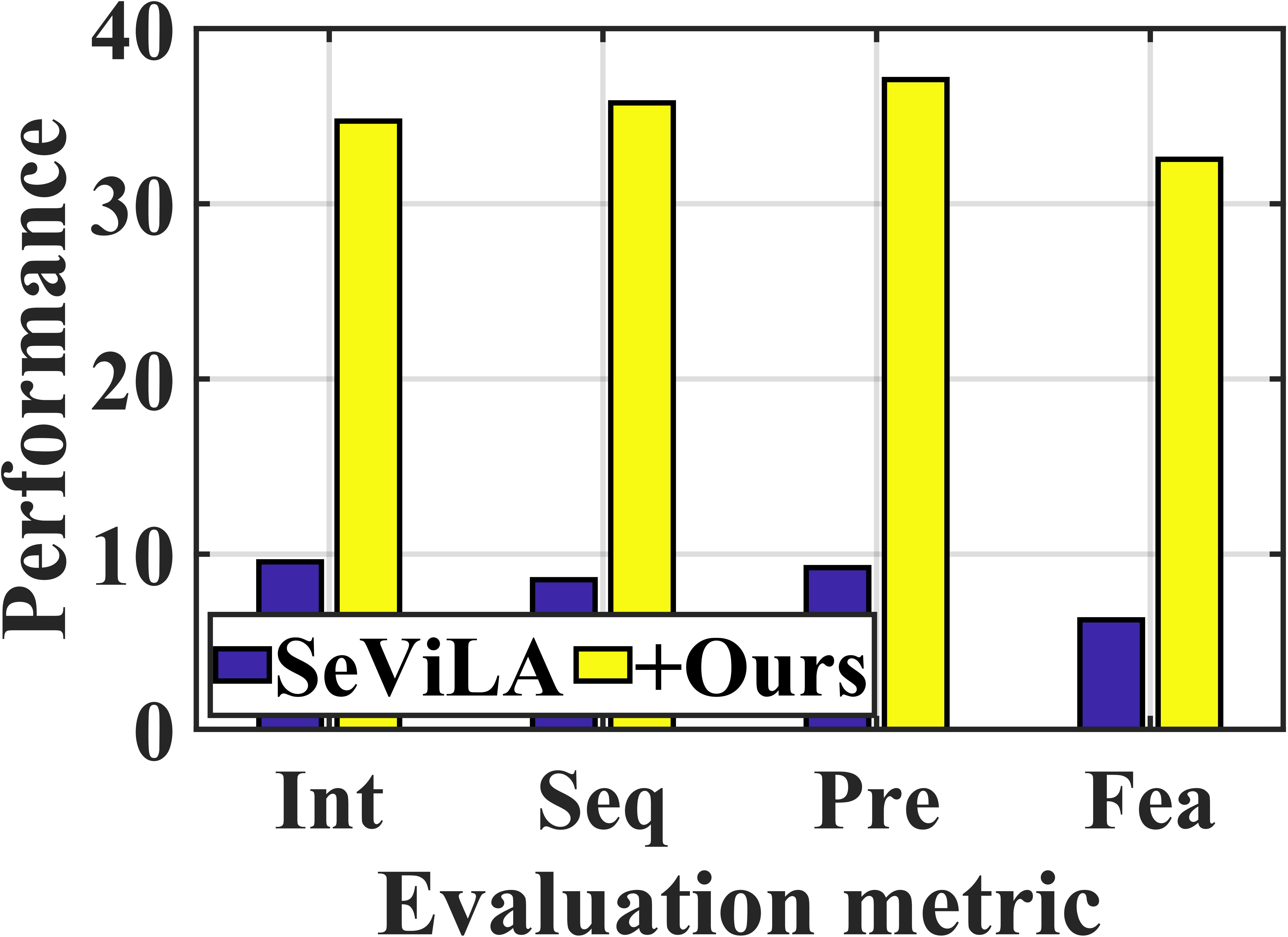} 
\hspace{-0.09in}
\includegraphics[width=0.16\textwidth]{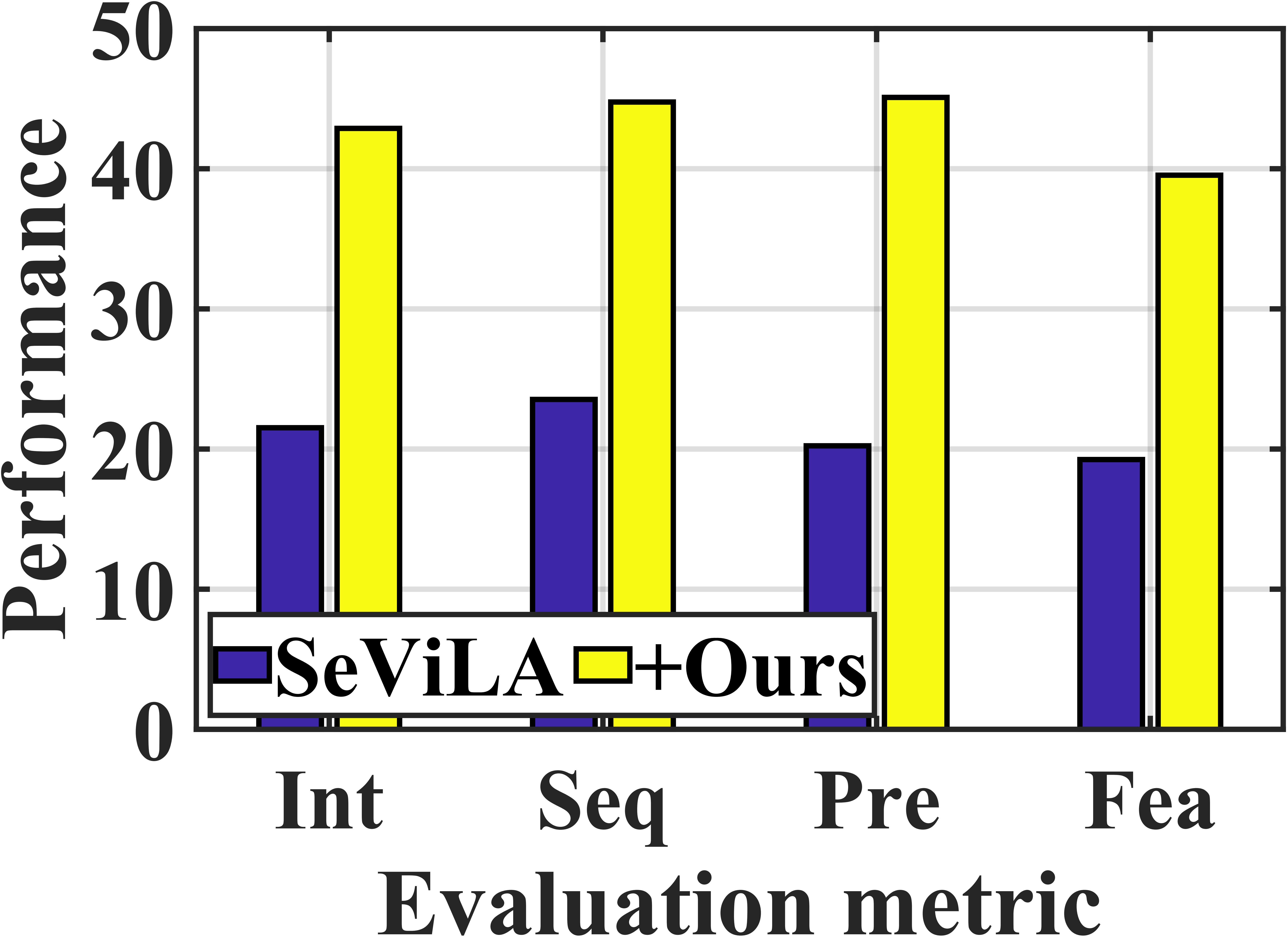} 
\caption{Balanced incomplete (left: incomplete(video) = incomplete(text) = 50\%) vs unbalanced incomplete (middle: incomplete(video) = 70\% and incomplete(text) = 30\%; right: incomplete(video) = 30\% and incomplete(text) = 70\%) for the ideoQA task on the STAR dataset.}
\label{fig:unbalance}
\end{figure}

\subsection{Performance Comparison}

For a fair comparison, we follow previous open-source  methods to directly cite the corresponding results from compared methods. Since our framework is a unified framework, we treat our framework as  the plug-and-play module for state-of-the-art  models to evaluate its effectiveness.


\noindent \textbf{Performance comparison on the VTR task.} 
In this task, we consider two significant subtasks: text-to-video  retrieval and video-to-text retrieval. Table \ref{tab:vtr} illustrates the effectiveness of our model as the plug-and-play module for previous VTR methods.
When inputting incomplete pairs,  all the compared methods suffer significant performance degradation.
This is mainly because these  VTR methods  ignore missing frames and words, and directly concatenate reserved words and frames for cross-modal alignment, which results in these methods not being able to fully understand the whole video and text. With our proposed framework as the plug-and-play module, these VTR methods can obtain significant performance improvement.


\noindent \textbf{Performance comparison on the VideoQA task.} 
Tables \ref{tab:nextqa} and \ref{tab:star} report the experimental results for  VideoQA, where the performance of previous methods can perform well on complete video-text pairs but suffer from severe performance degradation on incomplete video-text pairs.


\noindent \textbf{Performance comparison on the VSG task.} As for the VSG task, we adopt official train/test splits under both fully-supervised and weakly-supervised setting.
Table \ref{tab:vsg_act_cha}, Table \ref{tab:efficient} and Figure \ref{fig:tacos} summarize the quantitative comparison results. 
We can find that our proposed framework can serve as the plug-and-play module to effectively improve the performance of state-of-the-art VSG methods over all the metrics. The impressive performance of our framework illustrates its superiority. 


\noindent \textbf{Efficiency comparison.} 
To comprehensively evaluate our model, we compare the  efficiency and effectiveness of our framework (with 2D-TAN as the base model) with state-of-the-art open-source methods. 
In Table \ref{tab:efficient}, our model achieves much faster processing speeds with relatively fewer learnable parameters than most of these state-of-the-art methods. 

\noindent \textbf{Visualization.}
Figure~\ref{fig:vis} depicts the visualizations of three challenging tasks on various incomplete video-text datasets. It illustrates that state-of-the-art methods obtain poor performance on  incomplete multi-modal datasets. 


\subsection{Ablation Study and Analysis}

\noindent\textbf{Main ablation studies.} To evaluate the effectiveness of each module in our framework, we conduct ablation studies regarding the modules 
(\textit{i.e.}, Multi-modal Feature Approximation, Multi-modal Knowledge Distillation, Multi-granularity  Multi-modal Integration) in Table \ref{tab:main_ablation}.  In particular, we remove each key individual module while keeping the other  modules to investigate its contribution. For convenience, we design four ablation models: 1) Ours(a). We remove the  ``Multi-modal Feature Approximation'' module. 2) Ours(b). We remove the ``Generating Positive and Negative texts'' module. 3) Ours(c). We remove the ``Multi-granularity  Multi-modal Integration'' module. 
Besides, we treat our full model as the baseline: Ours(full).
In Table \ref{tab:main_ablation}, all the modules contribute a lot to the final performances on two challenging datasets, showing their effectiveness for   VSG. 

\noindent \textbf{Influence of the incompleteness rate.} To evaluate the influence of different incompleteness  rates on each modality,
we conduct ablation study  on the  LSMDC dataset. Figure~\ref{fig:missing_rate} illustrates the corresponding results.
For both video and text, as the miss rate increases, the performance of all the base methods drops severely. Fortunately, our framework can serve as the plug-and-play module for these methods to maintain the satisfactory performance.

\noindent \textbf{Balanced incompleteness vs unbalanced incompleteness.}
In Figure \ref{fig:unbalance}, we further evaluate the impact of the unbalanced incompleteness
, which illustrates the effectiveness of our framework since most of real-world datasets are unbalanced incomplete. 
 The satisfactory performance of our method is because our multi-modal feature approximation can mine the semantics from low-incompleteness modality to approximate the features of high-incompleteness for better multi-modal integration.

 \section{Conclusion}
In this paper, we target a new task: incomplete video-text alignment. A unified completeness network is proposed to address the modality-incomplete challenges in various downstream multi-modal tasks. 
We make the first attempt to pose a brand-new and realistic setting, incomplete video-text alignment for unbalanced incomplete multi-modal inputs. We propose a unified completeness network to address the modality-incomplete challenges in various downstream multi-modal tasks (video-text retrieval, video question answering and video sentence grounding).
Extensive experiments on many incomplete multi-modal datasets show that our framework can serve as the plug-and-play module for state-of-the-art VLM-based works to improve their performance.

{
    \bibliography{main}
 }


\end{document}